\documentclass[fleqn,10pt]{wlscirep}
\usepackage{graphicx} 
\usepackage[utf8]{inputenc}
\usepackage{multirow,amsthm,amssymb,amsbsy,amsmath,epsfig,epstopdf,bm}
\usepackage{subcaption}
\usepackage{color}
\usepackage{hyperref}
\usepackage{ulem} 
\newcommand{\R}{\mathbb{R}}
\newcommand{\Z}{\mathbb{Z}}

\title{A tensor network approach for chaotic time series prediction}

\author[1,*]{Rodrigo Mart\'inez-Pe\~na}
\author[1,2,3]{Rom\'an Or\'us}
\affil[1]{Donostia International Physics Center, Paseo Manuel de Lardizabal 4, E-20018 San Sebasti\'an, Spain}
\affil[2]{Multiverse Computing, Paseo de Miram\'on, E-20014 San Sebasti\'an, Spain}
\affil[3]{Ikerbasque Foundation for Science, Maria Diaz de Haro 3, E-48013 Bilbao, Spain}
\affil[*]{rodrigo.martinez@dipc.org}

\begin{abstract}
Making accurate predictions of chaotic time series is a complex challenge. Reservoir computing, a neuromorphic-inspired approach, has emerged as a powerful tool for this task. It exploits the memory and nonlinearity of dynamical systems without requiring extensive parameter tuning. However, selecting and optimizing reservoir architectures remains an open problem. Next-generation reservoir computing simplifies this problem by employing nonlinear vector autoregression based on truncated Volterra series, thereby reducing hyperparameter complexity. Nevertheless, the latter suffers from exponential parameter growth in terms of the maximum monomial degree. Tensor networks offer a promising solution to this issue by decomposing multidimensional arrays into low-dimensional structures, thus mitigating the curse of dimensionality. This paper explores the application of a previously proposed tensor network model for predicting chaotic time series, demonstrating its competitiveness in terms of accuracy and computational efficiency compared to conventional echo state networks. Using a state-of-the-art tensor network approach enables us to bridge the gap between the tensor network and reservoir computing communities, fostering advances in both fields. 
\end{abstract}
\begin{document}

\flushbottom
\maketitle

\thispagestyle{empty}

\section*{Introduction}

Chaos is a ubiquitous phenomenon in nature, and it is present in such diverse fields as weather and climate \cite{slingo2011uncertainty}, population dynamics \cite{may2007theoretical}, and nonlinear optics \cite{argyris2005chaos}. High sensitivity to initial conditions leads to exponentially growing errors and makes the efficient prediction of chaotic time series a challenging problem. A wide spectrum of data-driven machine learning models can be employed for this task, ranging from recurrent kernel methods \cite{gonon2022reservoir,grigoryeva2024infinite} to long
short-term memory networks \cite{vlachas2018data} and transformers \cite{gilpin2023model}. A popular approach in this direction is reservoir computing (RC) \cite{pathak2018model,griffith2019forecasting,vlachas2020backpropagation,grigoryeva2021chaos,grigoryeva2023learning}. Inspired by the field of neuromorphic computing, it exploits the intrinsic memory and nonlinearity of dynamical systems to process temporal information. The most important feature of this technique is that it does not require a fine-tuning of the dynamical system's parameters (usually randomly generated), relying only on the training of an (in most cases linear) output layer. This quick training, together with their universal approximation \cite{grigoryeva2018universal,grigoryeva2018echo,gonon2019reservoir,gonon2021fading} and embedding \cite{grigoryeva2021chaos,grigoryeva2023learning} properties, makes RC models very attractive for the forecasting of chaotic systems. 

Nevertheless, the selection and optimization of reservoirs is an open problem. A very popular approach is the echo state networks (ESNs) \cite{jaeger2002tutorial,lukovsevivcius2009reservoir,sun2022systematic,platt2022systematic}, which have demonstrated high performance in empirical studies and are also known for their universal approximation properties \cite{grigoryeva2018echo,gonon2019reservoir,gonon2021fading,gonon2023approximation}.  However, ESNs require the exploration of several hyperparameters for each task, which increases the complexity for real-world applications. An alternative strategy is the realization of dedicated hardware, which could enable speed gains and power savings with plenty of possible physical substrates \cite{tanaka2019recent,nakajima2020physical,nakajima2021reservoir}. However, although physical RC has the potential for industrial applications, several challenges prevent its large-scale adoption, and more research in both theoretical and experimental aspects\cite{yan2024emerging}. For instance,  postprocessing limits the overall computational speed of physical RC systems,  so hardware-based readouts are preferable to software-based ones \cite{antonik2016online,porte2021complete}.

A third approach is to change the reservoir by a nonlinear vector autoregression, where the state vector consists of a linear combination of delayed input monomials. This idea is known as next-generation reservoir computing (NGRC) \cite{gauthier2021next}, and is successful in predicting and controlling chaotic systems \cite{barbosa2022learning,gauthier2022learning,kent2024controllingChaos,kent2024controlling}. In this technique, it is possible to select the monomials to be introduced into the state vector, finding the maximum delay and maximum monomial degree as hyperparameters. The key feature behind NGRC is that state vectors are truncated versions of the Volterra series, which has universal approximation properties \cite{boyd1985fading,grigoryeva2019differentiable}. Still, state-space models given by a truncated Volterra series can be elusive due to the exponential growth of trainable parameters regarding the maximum monomial degree. Hence, there is a strong motivation to solve this curse of dimensionality, finding a promising direction in the field of tensor networks (TNs) \cite{batselier2017tensor,batselier2021enforcing,batselier2023khatri,khouaja2004identification,favier2009parametric,favier2023tensor}. 

Tensor networks are techniques based on decomposing multidimensional arrays into a set of lower-dimensional arrays, called core tensors. These core tensors can be efficiently manipulated and stored in memory, having applications in such diverse fields as quantum mechanics \cite{orus2019tensor,banuls2023tensor}, machine learning \cite{stoudenmire2016supervised,cheng2019tree,tomut2024compactifai}, and engineering \cite{favier2023tensor,batselier2022low}. This paper illustrates that a TN model can offer competitive performance and speed for chaotic time series prediction. In particular, we will implement one of the latest models proposed for truncated Volterra series \cite{batselier2023khatri}, whose hyperparameters are, similar to the NGRC case, the maximum delay and maximum monomial degree. 

We will compare the TN metrics against a conventional ESN model. To assess their performance, we use the $\texttt{dysts}$ database \cite{gilpin2021chaos,gilpin2023model}. This library provides a standardized collection of chaotic systems often used to benchmark time series forecasting methods. Our key goal is to bridge the gap between the TNs and RC communities, harnessing the developments of both fields during the last two decades.

The structure of the paper is as follows. The Section Methods introduces the notation, numerical details,  and the truncated Volterra series TN implementation of reference \citeonline{batselier2023khatri}. The Section Results contains the numerical results of benchmarking the models in terms of performance and computational speed. Finally, the Section Discussion summarizes the main contributions of the paper.

\section*{Methods}\label{sec:methods}

\subsection*{Tensors}

A $D$th-order tensor $\mathcal{X}\in\mathbb{R}^{I_1\times \dots \times I_D}$ is a multidimensional array with $D\in \mathbb{N}$ indices; that is, each entry can be written as $\mathcal{X}(i_1,\dots,i_D)$. Most common tensors are scalars ($D=0$), vectors ($D=1$), and matrices ($D=2$), and we will denote them as $x$, ${\bf x}$, and ${\bf X}$, respectively.  $D$th-order symmetric tensors $\mathcal{X}\in\mathbb{R}^{I\times \dots \times I}$, which are of particular interest for us, are those invariant under index permutation:
\begin{equation}
    \mathcal{X}(i_1,\dots,i_D) = \mathcal{X}(\pi(i_1,\dots,i_D)),
\end{equation}
where $\pi(i_1,\dots,i_D)$ stands for any possible permutation of $D$ indices. We denote by $\text{Sym}_D(\R)$ the vector space of symmetric tensors of order $D$ over the reals. 

Let us now introduce some important tensor transformations. The reshape operation modifies the order of a tensor such that a $D$th-order tensor $\mathcal{X}\in\mathbb{R}^{I_1\times \dots \times I_D}$ is rearranged to form a $\mathcal{Y}\in\mathbb{R}^{J_1\times \dots \times J_K}$ without changing the total number of entries, that is, $\prod^D_{d=1}I_d=\prod^K_{k=1}J_k$. We denote the operation as $\mathcal{Y}=\text{reshape}(\mathcal{X},\{J_1,J_2,\dots,J_K\})$. The vectorization is a reshape operation that converts a $D$th-order tensor $\mathcal{X}\in\mathbb{R}^{I_1\times \dots \times I_D}$ in a vector ${\bf x}\in\mathbb{R}^{\prod^D_{d=1}I_d}$, and we will denote it as ${\bf x}=\text{vec}(\mathcal{X})$. The conversion of $D$ indices $i_1,\dots,i_D$ into a multi-index $[i_1 i_2\dots i_D]$ is defined as
\begin{equation}
    [i_1 i_2\dots i_D]=i_1+\sum^D_{d=2}(i_d-1)\prod^{d-1}_{m=1}I_m,
\end{equation}
such that the entries of a vectorized tensor fulfill
\begin{equation}
    \mathcal{X}(i_1,\dots,i_D)={\bf x}([i_1 i_2\dots i_D]).
\end{equation}

In this manuscript, we are interested in efficient decompositions of matrices with the curse of dimensionality. A well-established technique for matrix decomposition is low-rank approximation, with singular value decomposition (SVD) being a prominent example \cite{horn2012matrix}. The rank of a matrix, denoted as $R$, corresponds to the number of its nonzero singular values. By truncating small singular values, we can enhance storage efficiency and computational performance while maintaining an effective approximation of the original matrix \cite{kalman1996singularly}. This concept extends beyond matrices to higher-order tensors, giving rise to various tensor decomposition methods, each with distinct characteristics and rank definitions \cite{oseledets2009breaking,kolda2009tensor,oseledets2011tensor}. 

In this work, we utilize matrix product operators (MPOs) for matrix decomposition. An MPO, also known as a tensor train matrix, decomposes a matrix with exponential dimensions in a linear sequence of core tensors \cite{verstraete2004matrix,zwolak2004mixed,oseledets2010approximation}. Consider a matrix ${\bf X}\in \mathbb{R}^{I^D\times J^D}$. We split each row and column into $D$ indices $i_1,\dots, i_D$ and $j_1,\dots,j_D$, defining the multi-indices $[i_1 i_2\dots i_D]=i_1+\sum^D_{d=2}(i_d-1)I^{d-1}$ and  $[j_1 j_2\dots j_D]=j_1+\sum^D_{d=2}(j_d-1)J^{d-1}$, such that ${\bf X}$ can be reshaped into a 2$D$th-order tensor $\mathcal{X}\in\mathbb{R}^{I\times \dots \times I\times J\times \cdots \times J}$. This tensor can be decomposed as an MPO that consists of $D$ fourth-order tensors $\mathcal{X}^{(d)}\in \mathbb{R}^{R_d\times i_d\times j_d\times R_{d+1}}$, such that each matrix entry can be computed as
\begin{equation}
    {\bf X}([i_1\dots i_D],[j_1\dots j_D])=\sum^{R_1}_{r_1=1}\dots \sum^{R_D}_{r_D=1}\mathcal{X}^{(1)}(r_1,i_1,j_1,r_2)\mathcal{X}^{(2)}(r_2,i_2,j_2,r_3)\dots \mathcal{X}^{(D)}(r_D,i_D,j_D,r_1),
\end{equation}
where $R_1 = 1$.

\subsection*{Truncated Volterra series}
Let us introduce the truncated Volterra series. Let $M\in \mathbb{N}$ be the maximum delay of the series and $D\in \mathbb{N}$ the maximum order of the monomials. We define each $d$th-order Volterra series kernel $h_d(m_1,\dots,m_d)$ as a function of the delays $m_j$ for $j=1,\dots , d$ and $m_j=0,\dots,M-1$. Let input and output be bi-infinite discrete-time sequences of the form $\underbar{\bf u}=(\dots, u(-1),u(0),u(1),\dots)\in \R^\Z$ and $\underbar{\bf y}=(\dots, y(-1),y(0),y(1),\dots)\in \R^\Z$ respectively. Then, the single-input single-output truncated series is defined as
\begin{equation}
    y(n) = h_0 +\sum^D_{d=1}\sum^{M-1}_{m_1,\dots,m_d = 0} h_d(m_1,\dots,m_d)\prod^d_{j=1} u(n-m_j).
\end{equation}
Now, following e.g., reference \citeonline{batselier2017tensor}, we will define the multivariate case in a compact matrix form. Let us define the $P$-dimensional input and $L$-dimensional output as
\begin{equation}\begin{split}
    &\textbf{u}(n) := \begin{pmatrix}
        u_1(n), u_2(n), \dots, u_p(n)
    \end{pmatrix}^\top\in \mathbb{R}^P,    \\
    &\textbf{y}(n) := \begin{pmatrix}
        y_1(n),y_2(n),\dots, y_p(n)
    \end{pmatrix}^\top\in \mathbb{R}^L,
\end{split}
\end{equation}
for all $n\in \Z$. To obtain the truncated Volterra series, one needs to compute all the possible monomials of the input in terms of the hyperparameters $M$ and $D$. We start by introducing an extended vector of the input as
\begin{equation}
    \textbf{u}_n := \begin{pmatrix}
        1,\textbf{u}(n)^\top, \cdots, \textbf{u}(n-M+1)^\top
    \end{pmatrix}^\top\in \mathbb{R}^{I},  
\end{equation}
with $I:=PM+1$. Then, we define the vector containing all the monomials of the input up to degree $D$ using repeated Kronecker products:
\begin{equation}
   \textbf{u}^D_n:= \overbrace{\textbf{u}_n \otimes \cdots \otimes \textbf{u}_n }^{D \text{ times}}\in \mathbb{R}^{I^D}.
\end{equation}
The $\textbf{u}^D_n$ vectors are a vectorization of a symmetric tensor of $D$ indices with $I$ possible values each. This fact will be relevant for the next section. Then, at each time step, the truncated Volterra series is determined by
\begin{equation}
    \textbf{y}(n)^\top = \left( \textbf{u}^D_n\right)^\top \textbf{H},\quad n\in \Z,
\end{equation}
where $\textbf{H}$ is the Volterra coefficients matrix with dimension $I^D \times L$. This equation can be extended to include $N$ time steps as follows:
\begin{equation} 
    \begin{pmatrix}
    \textbf{y}(1)^\top\\
    \vdots\\
    \textbf{y}(N)^\top
    \end{pmatrix}=\begin{pmatrix}
        \textbf{u}^D_1 & \cdots & \textbf{u}^D_N
    \end{pmatrix}^\top \textbf{H}, 
\end{equation}
or in a more compact form,
\begin{equation} \label{eq:Y=UH}
 \textbf{Y} = \textbf{U}\textbf{H},
\end{equation}
where matrices $\textbf{Y}$ and $\textbf{U}$ have dimensions $N\times L$ and $N\times I^D$ respectively. Since each row of matrix $\textbf{U}$ is given by the $D$-times repeated Kronecker product, we can rewrite $\textbf{U}$ in terms of repeated row-wise Khatri-Rao products \cite{batselier2018tensor}. Define the $N\times I$ matrix $\tilde{\textbf{U}}$ as 
\begin{equation}
    \tilde{\textbf{U}} := \begin{pmatrix}
        \textbf{u}^\top_1 \\
        \vdots \\
        \textbf{u}^\top_N
    \end{pmatrix},
\end{equation}
then \textbf{U} can be rewritten as
\begin{equation}
    \textbf{U} = 
    \overbrace{\tilde{\textbf{U}}\odot \cdots \odot \tilde{\textbf{U}} }^{D \text{ times}},
\end{equation}
where $\odot$ denotes the row-wise Khatri-Rao product.

Finally, we define the learning problem: Given a finite sequence $\left\{\textbf{u}(n),\textbf{y}(n)\right\}^N_{n=1}$ of training data, a maximum monomial degree $D$ and a maximum delay $M$, find the unknown $\textbf{H}$ matrix that better fits the training dataset. Since Eq.~\eqref{eq:Y=UH} is a linear system, the naive solution would be to go straightforward for the least squares solution, given by
\begin{equation}\label{eq:H=U+Y}
    \textbf{H} = \textbf{U}^+\textbf{Y},
\end{equation}
with $\textbf{U}^+$ being the Moore–Penrose pseudoinverse. However, the exponential scaling in terms of the hyperparameter $D$ makes this computation hard in general. Some solutions to solve this curse of dimensionality are: PARAFAC \cite{khouaja2004identification,favier2009parametric,favier2012nonlinear,bouilloc2012nonlinear,favier2023tensor}, MPOs optimized with DMRG \cite{batselier2017tensor}, MPOs with extra structure using a Bayesian framework \cite{memmel2023bayesian}, and MPOs enforcing symmetry in the tensors’ solution
\cite{batselier2021enforcing,batselier2023khatri}. We will focus on the latter case \cite{batselier2023khatri}, since it is the simplest algorithm in terms of the number of hyperparameters and it offers high speed of training. Nevertheless, our source code also includes the implementation of references \citeonline{batselier2017tensor,batselier2021enforcing}, which can be of interest to some readers.

\subsection*{MPOs enforcing symmetry}
Each row of the matrix $\textbf{U}$ in Eq.~\eqref{eq:Y=UH} is the vectorization of a symmetric tensor. This means that the vectors $\textbf{u}^D_n$ contain repeated entries due to the symmetry introduced by the Kronecker products. This symmetry bounds the rank of $\textbf{U}$, as shown in Lemma 4.2 of \citeonline{batselier2017tensor}, namely:
\begin{equation}
    \text{rank}(\textbf{U}) \leq R = \binom{PM+D}{PM}.
\end{equation}
where $R$ is the maximum number of independent coefficients (or monomials) of the truncated Volterra series.

Since matrix $\textbf{U}$ has dimensions $N\times I^D$, it is a rank-deficient matrix and the normal equation 
\begin{equation}
    {\bf U}^\top{\bf U}{\bf H}= {\bf U}^\top{\bf Y} 
\end{equation}
for the least squares problem of Eq.~\eqref{eq:Y=UH} has an infinite number of solutions, independently of the number of samples $N$. In Proposition 4.2 of \citeonline{batselier2017tensor}, it was shown that under the assumption of persistently exciting input, that is, $\text{rank}({\bf U}) = R$ ($N\geq R$), each column of the unique minimal-norm solution for $\textbf{H}$ is the vectorization of a $D$-dimensional symmetric tensor. We show in the Appendix that this result is independent of the rank of ${\bf U}$. Therefore, enforcing symmetry on the columns of $\textbf{H}$ is equivalent to finding the minimum-norm solution of Eq.~\eqref{eq:Y=UH}, which is an implicit form of training regularization.

The minimum-norm solution is computed using the pseudoinverse, which is obtained through SVD. Assume that $N>R$, the SVD of matrix $\textbf{U}$ is
\begin{equation}\label{eq:thinSVD}
    \textbf{U} = \textbf{Q}\begin{pmatrix}
        \textbf{S} & 0\\
        0 & 0 \\
    \end{pmatrix}\textbf{V}^\top=  \begin{pmatrix}
        \textbf{Q}_1 & \textbf{Q}_2
    \end{pmatrix}
    \begin{pmatrix}
        \textbf{S} & 0\\
        0 & 0 \\
    \end{pmatrix}\begin{pmatrix}
        \textbf{V}^\top_1 \\
        \textbf{V}^\top_2  
    \end{pmatrix} =   \textbf{Q}_1 \textbf{S}\textbf{V}^\top_1,
\end{equation}
where $\textbf{Q}\in \mathbb{R}^{N\times N}$, $\textbf{V}\in \mathbb{R}^{I^D\times I^D}$, $\textbf{Q}_1\in \mathbb{R}^{N\times R}$, $\textbf{V}_1\in \mathbb{R}^{I^D\times R}$ are orthogonal matrices and $S\in \mathbb{R}^{R\times R}$ is the diagonal matrix of non-zero singular values. Finally, the symmetric Volterra kernel coefficients can be obtained as
\begin{equation}\label{eq:Hsymm_solution}
    \textbf{H}=  \textbf{V}_1 \textbf{S}^{-1}\textbf{Q}^\top_1\textbf{Y}.
\end{equation}
 Computing the SVD of $\textbf{U}$ is computationally inefficient due to the exponential number of columns in terms of $D$. However, this problem can be overcome if MPOs are introduced to decompose both $\textbf{U}$ and $\textbf{H}$ in Eq.~\eqref{eq:Y=UH}. The construction of these MPOs was originally proposed in \citeonline{batselier2021enforcing}, where repeated Khatri-Rao products together with truncated SVDs were required. However, reference \citeonline{batselier2023khatri} proposes to replace this construction with an exact representation that does not involve computations and only instances of matrix $\tilde{\textbf{U}}$ need to be stored. We refer to \citeonline{batselier2023khatri} for further details of this construction.

\subsection*{ESN benchmark}
Echo state networks (ESNs) were introduced two decades ago as an alternative to the more traditional recurrent neural networks (RNNs) \cite{jaeger2001echo}. ESNs have become one of the most popular RC families due to their simplicity, easy training, and competitive performance compared with RNNs. In this work, we use ESNs as a benchmark to compare with the performance and training time of the TN model. 

Let us define $\textbf{x}(n)\in \mathbb{R}^{N_r}$ as the state vector for $N_r$ neurons at time step $n\in \Z$, its dynamics is given by
\begin{equation}
      \textbf{x}(n) = \tanh (W\textbf{x}(n-1)+g\textbf{v}\textbf{u}(n)),
\end{equation}
where $W\in \mathbb{R}^{N_r\times N_r}$ is the connectivity matrix, $\textbf{v}\in \mathbb{R}^{N_r\times P}$ is the connectivity input vector, $g\in \mathbb{R}^+$ is the input strength, and $\tanh(\cdot)$ is applied element-wise. The elements of arrays $W$ and $\textbf{v}$ are randomly generated from a uniform distribution $[-1,1]$, while the spectral radius $\rho$ of the matrix $W$ will be controlled as a hyperparameter.

The output layer ${\bf y}(n)\in \mathbb{R}^{L}$ of the ESN model is constructed as follows: 
\begin{equation}
    {\bf y}(n)^\top = {\bf x}(n)^\top W^{\text{out}}+{\bf b}^\top,
\end{equation}
where $W^{\text{out}}\in \mathbb{R}^{N_r\times L}$ is the output weight matrix and ${\bf b}\in \R^L$ is a constant offset. Given a training set of length $N$, the linear inversion problem can be defined by 
\begin{equation}
  \textbf{Y} = \textbf{X}W^{\text{out}}+{\bf 1}_N{\bf b}^\top,
\end{equation}
where $ \textbf{Y}\in\mathbb{R}^{N\times L} $, $\textbf{X}\in \mathbb{R}^{N\times N_r}$, and ${\bf 1}_N:=(1,\dots,1)^\top\in \R^{N}$. We use Tikhonov regularization, finding the following ridge regression solution:
\begin{equation}
    \hat{W}^{\text{out}} = ({\bf X}^\top{\bf A}_{N}{\bf X}+\lambda {\bf I}_{N_r})^{-1}{\bf X}^\top {\bf A}_{N}{\bf Y},\quad \hat{\bf b}^\top = \frac{1}{N}{\bf 1}^\top_N(\textbf{Y}-\textbf{X}\hat{W}^{\text{out}}),
\end{equation}
where ${\bf I}_{N_r}\in \R^{N_r\times N_r}$ is the identity matrix, ${\bf A}_{N}:={\bf I}_N-{\bf 1}_N{\bf 1}^\top_N/N$, and $\lambda>0$ is the regularization hyperparameter.

\subsection*{Dataset}

Our target tasks are low-dimensional chaotic dynamical systems from the $\texttt{dysts}$ database \cite{gilpin2021chaos,gilpin2023model}. Introduced in 2021, this library standardizes the chaotic systems that are frequently used for benchmarking time series forecasting methods. The database contains data, equations, and dynamic properties for more than 100 chaotic systems with strange attractors from various scientific disciplines. To consistently choose integration and sampling timescales, this database computes the power spectrum for all systems to identify both the smallest and dominant significant frequencies, where significance is determined relative to random phase surrogates \cite{gilpin2023model}. The smallest significant frequency is identified as the optimal integration step. Once integrated, the dominant significant frequency $1/T$, called the period, is used to resample the trajectory, using 100 points per period in our case, i.e., a sampling rate of $\Delta t = T/100$. 

To simplify the hyperparameter optimization of the TN model, we follow the approach of \citeonline{kaptanoglu2023benchmarking} and select 70 target tasks where the ODEs have polynomial nonlinearities with degree no more than four. Also, the phase space dimension of these chaotic systems is either $P=3$ or $P=4$. 

Trajectories are generated with a slightly modified version of the code of reference \citeonline{kaptanoglu2023benchmarking}. We generate two trajectories of 21001 points for each dynamical system (or task). The first trajectory is used for training and validation, while the second is for training and testing. Each trajectory contains windows made of three pieces: $N_{\text{warmup}}=5000$ points for washing out the initial conditions of the ESN model (which we truncate for the TN model so we train and evaluate over the same data), $N_{\text{train}}=10000$ points for training, and $N_{\text{val}} =N_{\text{test}}=5000$ points for validation and testing. The 1000 extra points of the trajectories are used for rolling a window during cross-validation and testing, as will be explained below. The trajectories are normalized with respect to the maximum and minimum values of the training set, which lies in the interval $[0, 1]$. This ensures that no information from the validation or test sets is transferred to the training set.

We will address two different types of problems. In the \textit{memoryless} case, all dynamical variables are used as inputs/outputs for training, validation, and testing, with $P=3$ or $P=4$ depending on the dynamical system. We call it the memoryless case because the discretization of the integration method used to generate the trajectories depends only on one time step in the past. Then, given the state of the discretized dynamical system, no further delayed states are required to predict the next time step. In the \textit{memory} case, only the first dynamical variable is used as input/output for training, validation, and testing, with $P=1$ for all dynamical systems. In this problem of dynamical reconstruction from a single variable observation, Takens' Embedding Theorem guarantees that $2S+1$ delayed inputs suffice to learn the attractor dynamics \cite{takens2006detecting}, where $S$ is the dimension of the manifold on which the dynamical system is defined, although a lower bound could work depending on the dimension of the strange attractor \cite{sauer1991embedology}. 

\subsection*{Validation}
The hyperparameters will be selected using a rolling window method with cross-validation for time series. We split the training-validation trajectories into 5 windows, each one with $N_{\text{window}}=1+N_{\text{warmup}}+N_{\text{train}}+N_{\text{val}}=20001$ points, whose initial points are equidistant. We calculate the average validation metric for each set of hyperparameters over the five windows, then select the set with the lowest value as a grid search. Other approaches could potentially be explored, such as Bayesian optimization \cite{platt2022systematic}. 

For the TN model, we will evaluate $D\in\{2,3,4\}$ as the maximum monomial degrees, according to the nonlinearity of the chosen target tasks. For maximum delay, we will split into two cases: $M\in\{1,2,3,4\}$ for the memoryless case, and $M\in\{1,2,3,4,5,6,7,8,9,10\}$ for the memory case. These sets of $M$-values guarantee the persistently exciting input condition $N_{\text{train}}>R$, having $R=4845$, for $M=D=P=4$, and $R=1001$ for $P= 1$, $M = 10$, and $D = 4$.
  
For the ESN model, we fix the input strength to $g = 1$ and set $N_r=300$ for the memoryless case and $N_r = 500$ for the memory case.  
For validation, we explore the following sets for the spectral radius and the regularization hyperparameter: $\rho \in [0.0,0.1,0.2,\dots,1.5]$, and $\lambda\in [10^{-13},10^{-12},10^{-11},\dots,10^{-1}]$. One could still perform a more exhaustive exploration of the ESN hyperparameter space. However, we expect that since ESN models can be approximated by a truncated Volterra series, we could always find a TN model that performs as well as the ESN while requiring a much simpler hyperparameter exploration.

To validate the performance, the models are run autonomously; after training, the outputs are fed back as inputs to predict the next time step. Two different validation metrics will be explored: one focusing on long-term predictions --- the Wasserstein distance --- and another focusing on short-term predictions --- the normalized mean square error (NMSE). 

The Wasserstein distance measures the similarity between two probability distributions, here the spectra of two temporal series. For one-dimensional signals, it is known \cite{bagge2020new} that the  Wasserstein $p$-distance ($p\geq 1$) is given by the following integral: 
\begin{equation} \begin{split}
       & W_p (\Phi_1,\Phi_2) = \int^1_0 |F_1^{-1}(x)-F_2^{-1}(x)|^pdx,
\end{split}
\end{equation}
where $\Phi_1,\Phi_2$ are the probability distributions and $F_1^{-1},F_2^{-1}$ are the inverses of the cumulative distributions of $\Phi_1$ and $\Phi_2$, respectively. Numerically, we calculate $W_2$ (for $p=2$) with the function \texttt{WelchOptimalTransportDistance} from the library \texttt{SpectralDistances.jl} \cite{bagge2020new}. It computes the distance between two spectra estimated using a traditional spectral-estimation technique, the Welch method. For the memoryless case, our final validation metric is the average of this distance over all the phase space variables. 

The NMSE metric measures the mean square Euclidean distance between two sequences of data, normalized by the mean square Euclidean norm of the target data:
\begin{equation}
    NMSE = \frac{\sum^n_{j=1}\|{\bf y}(j)-\hat{\bf y}(j) \|^2_2}{\sum^n_{j=1}\|\hat{\bf y}(j) \|^2_2},
\end{equation}
where ${\bf y}(j)$ and $\hat{\bf y}(j)$ are prediction and target respectively.

\subsection*{Test metrics}
For testing the models, we will evaluate two different aspects: their short-term memory and long-term memory (or climate) forecasting capabilities. To quantify the short-term prediction capabilities, we compute the valid prediction time (VPT) as in \citeonline{wikner2024stabilizing}, which quantifies the number of Lyapunov times that the model predicts under a certain precision. The Lyapunov time is the timescale for which a chaotic dynamical system is predictable, and it is computed as the inverse of the maximum Lyapunov exponent of the task, $\Lambda$. The VPT expression is the following:
\begin{equation}
    \text{VPT} = \Lambda\Delta t\left(\min_{N_{\text{resyn}}< n\leq N_{\text{resyn}}+N_{\text{pred}}}\left\{n \ | \  \frac{\|{\bf y}(n)-\hat{\bf y}(n) \|_2}{\overline{E}}>\delta\right\}-N_{\text{resyn}}-1\right),
\end{equation}
where $N_{\text{resyn}}$ is the number of washout inputs to resynchronize the trained model, $\Delta t$ is the sampling time step of the task, $\delta$ is the valid time error threshold, and $\overline{E}$ is the average distance between target states computed from the initial training and warming up data as the mean of $\| \hat{\bf y}(j)-\hat{\bf y}(k) \|$ over $1 \leq j < k \leq N_{\text{warmup}}+N_{\text{train}}$ to normalize. We take $\delta = 0.2$, which represents a $20\%$ error in the prediction, and we define $N_{\text{pred}}=4000$, which is large enough for all tasks. 

The VPT metric is averaged over 100 different initial conditions. These initial conditions are obtained by displacing the initial point of the test set evenly with a rolling window. To resynchronize the ESN model, we fix $N_{\text{resyn}} = 5000$, also truncating these points for the TN model. 

For long-term memory, we compute the Wasserstein distance as we did for validation, obtaining the difference between the target and predicted spectra. We used $N_{\text{test}} = 5000$ points with no average over different initial conditions. To have a reference value for what constitutes a bad prediction, we compute the average Wasserstein distance over 50 trajectories with random initial conditions on the attractor.

These testing metrics are obtained over new trajectories, where we train the ESN and TN models with the optimized hyperparameters using the same division as before, $N_{\text{warmup}} = 5000$ and $N_{\text{train}} = 10000$. 

\section*{Results}\label{sec:results}
The prediction capabilities of the TN and ESN models are demonstrated with the test metrics. Most numerical computations were done on a laptop with 16 GB of RAM and an Intel Core i7-13620H CPU. The optimization of the ESN and TN models took longer ($\sim 22$h and $\sim 51$h for the longest ones, respectively), so these codes were run on a cluster. The code is available at \url{https://github.com/RMPhys/tnrc_paper}.
\begin{figure*}[h!]
\captionsetup[subfigure]{}
\begin{center}
\includegraphics[scale=0.6]{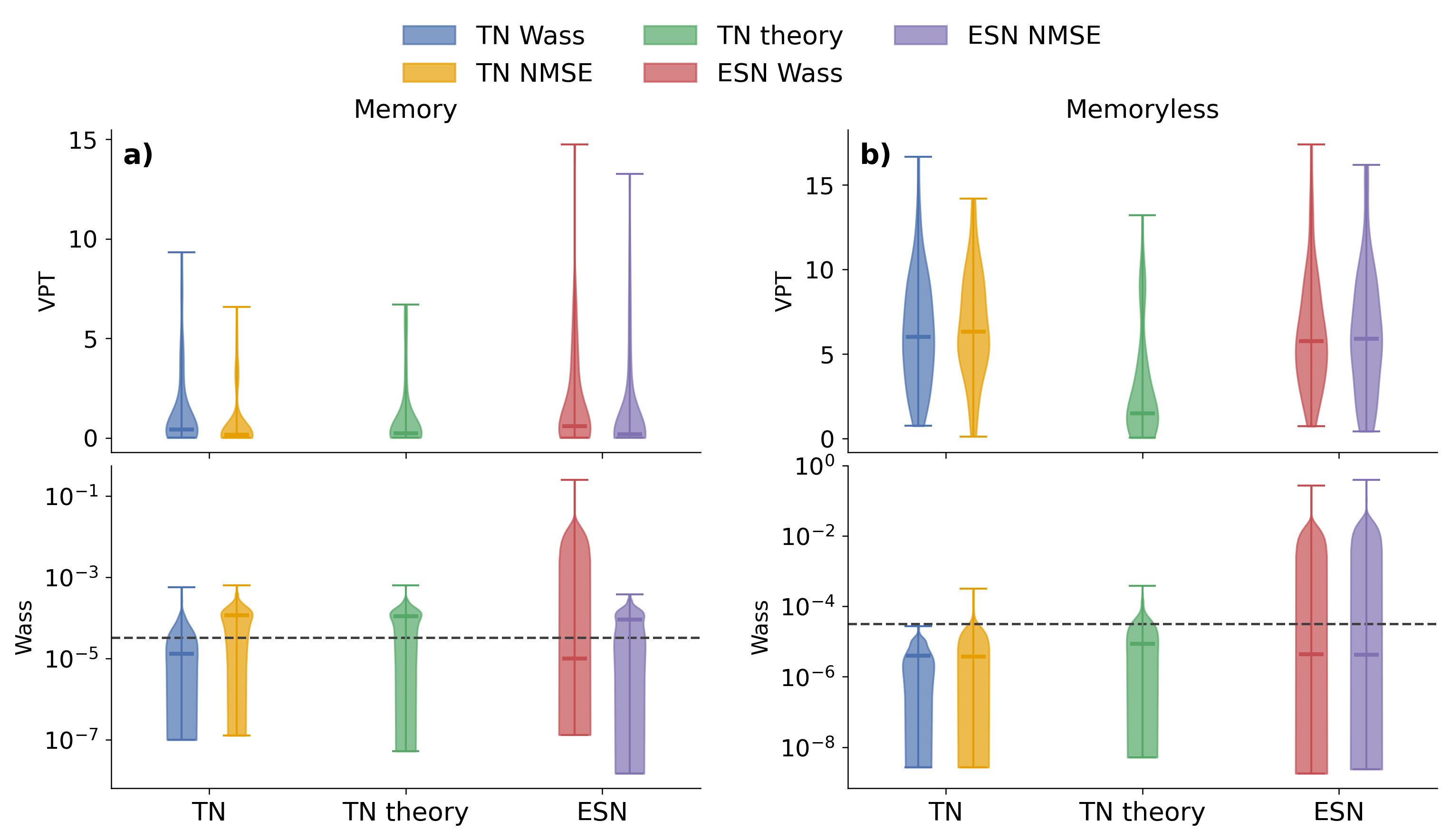}
\caption{Comparison of the VPT and Wasserstein metrics for the TN and ESN models.}\label{fig1}
\end{center}
\end{figure*}

Figure \ref{fig1} shows the statistics of the VPT and Wasserstein metrics for the 70 target tasks in both a) the memory and b) memoryless cases. Results for each specific task are shown in Appendix A in Fig.~\ref{fig3}. The width of the violins represents the distribution of values, while the inner line is the median, and the highest and lowest lines represent the maximum and minimum, respectively. For each plot, we display the TN and ESN models optimized using the Wasserstein distance (in blue and red), the models optimized using the NMSE metric (in orange and purple), and the TN model for which the hyperparameters are not optimized but are instead determined by the required nonlinearity and memory (in green). The required degree of nonlinearity is obtained by computing the maximum nonlinear monomial of the differential equations. The required memory depends on the case. For memoryless tasks, we set $M = 1$, whereas for memory tasks, we take $M = 7$ for 3-dimensional systems and $M = 9$ for the 4-dimensional ones. Finally, the gray dashed line in the Wasserstein plots represents the median of the 50 random surrogates.

We observe three things in Fig.~\ref{fig1}. Firstly, the TN theory model with the required hyperparameters performs statistically worse than the optimized TN model. Regarding the memoryless case, where we set $M=1$, other studies have also observed that a bit of memory can help with the prediction of chaotic time series, such as when studying ESNs near $\rho=0$ \cite{haluszczynski2019good,hart2023estimating,jaurigue2024chaotic}. For the memory case, the Takens' upper bound on $M$ does not guarantee it to be the optimal one. Altogether, these results indicate that optimization is also relevant for the TN model to ensure competitive performance.

Secondly, optimization using the Wasserstein metric produces slightly better results for VPT in a), while significantly improving the climate test compared to NMSE optimization. However, in b), the NMSE metric slightly improves the median value for VPT, while decreasing the maximum and minimum VPT values and maintaining the same climate test (or achieving an even worse maximum value). In other words, optimization using a long-term metric appears to accurately capture the memory requirements for VPT and climate predictions, whereas optimization using a short-term metric only marginally improves the median VPT of the memoryless case. These results suggest that using climate metrics for hyperparameter optimization is a promising approach, as recently discussed in \citeonline{fumagalli2025data}.

Thirdly, although the median values of the optimized models are similar when using the same optimization metric, the TN models have shorter distributions for both the VPT and Wasserstein metrics. For VPT results, the ESN models reach larger values than the TN models for some tasks (though not by much in b)), while the ESN models can also achieve much worse results than the TN models for the Wasserstein metric. This smaller variability could make TN models more appealing for simultaneously learning several chaotic dynamics. 

\begin{figure}[h!]
\captionsetup[subfigure]{}
\begin{center}
\includegraphics[scale=0.6]{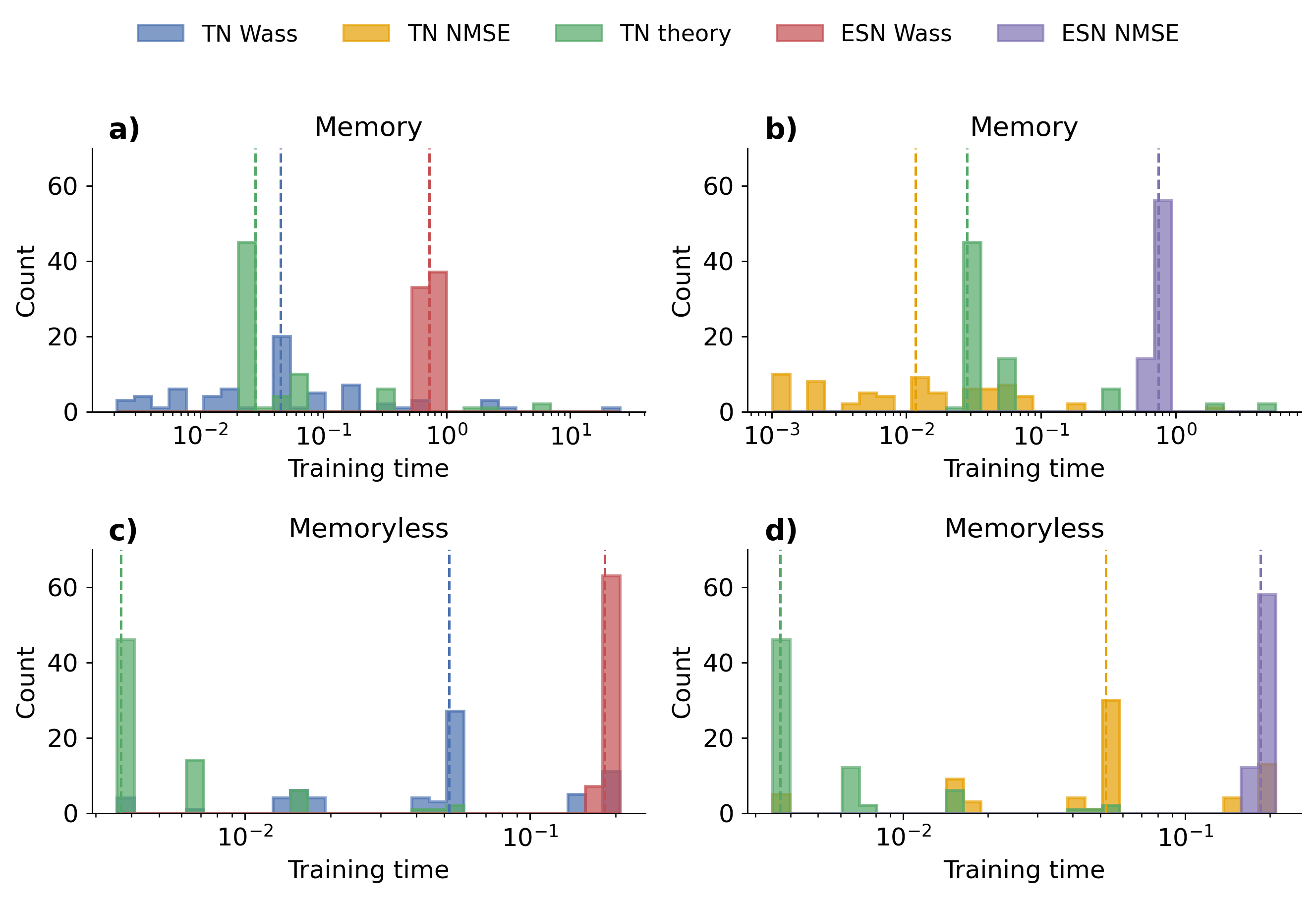}
\caption{Histograms of CPU training times in seconds for the optimized models.}\label{fig2}
\end{center}
\end{figure}

Next, we compare the distribution of training times in Fig.~\ref{fig2}. We remark that for ESN models, the training time includes the washout phase --- which is intrinsic to dynamical RC models ---, state matrix generation and linear regression. For TN models, the training time consists of the entire MPO construction,  and we skip the washout phase by truncating. Figure \ref{fig2} displays four histograms, where the x-axis shows the CPU training time for one thread in seconds and the y-axis shows the number of tasks. The vertical dashed lines correspond to the median value of the distributions. Figures a) and b) show the memory tasks and figures c) and d) show the memoryless tasks. To better visualize the results, the first column shows the TN and ESN models optimized with the Wasserstein metric, while the second column displays the models optimized with the NMSE metric. The TN model with the required hyperparameters (TN theory) is included in all plots for the sake of comparison. 

On the one hand, we observe that ESN models have very narrow distributions of training times. ESN hyperparameters do not significantly affect the running time of the dynamics, with very similar training times observed between a) and b), and c) and d). However, note the difference in training time between the first and second rows, which is almost one order of magnitude. This is because $N = 500$ and $P = 1$ were used for memory tasks (first row), and $N = 300$ and $P \in \{3, 4\}$ were used for memoryless tasks (second row). 

On the other hand, TN models occupy wider ranges. The left part of the histograms is populated by low values of $P$, $M$ and $D$, while larger values can result in training times that are orders of magnitude longer, especially in the memory case where larger values of $M$ are explored. These wide ranges are explained by the computational complexity of the algorithm. The most expensive step is an SVD computation with a complexity of $O(N^2_{\text{train}}IR_D)$ \cite{batselier2023khatri}, where \begin{equation}
    R_D=\binom{PM+D-1}{PM}=\binom{PM+D-1}{D-1}=\frac{1}{(PM)!}\prod^{PM-1}_{j=0}(D+j) = \frac{1}{(D-1)!}\prod^{D-1}_{j=1}(PM+j).
\end{equation}
When $PM$ is constant, $R_D$ grows with $O(D^{PM})$, while when $D$ is constant, the complexity of $R_D$ is $O((PM)^{D-1})$. Since we are exploring low values of $D$ --- with $D\in\{2,3,4\}$ --- against large values of $PM$ --- with the largest being $PM=10$ for the memory case and $PM=16$ for the memoryless case ---, we can assume that the leading growth is $O((PM)^{D-1})$. In this case, the SVD computation complexity would grow in a polynomial fashion, i.e. $O(N^2_{\text{train}}(PM)^{D})$. 

For the particular case of the TN theory model (green), we observe clearly defined decreasing columns. This decay in training times is due to the frequency of maximum nonlinearity in the target tasks. The most frequent value is $D = 2$, which, when combined with the lowest value of $M$ ($M = 7$ for the first row and $M = 1$ for the second row), corresponds to the largest column. See the notebook \texttt{plots.ipynb} in the source code for further details of all column compositions.


Finally, the optimal TN models generally take less time to train than the optimal ESN models. The median values of the optimal TN models are below the median ESN training times, with the majority of counts falling below the ESN bars. For memory tasks, the difference is more than one order of magnitude; for memoryless tasks, it is slightly smaller. However, a few TN model training times are larger than those of the ESNs. Notably, for memory tasks, TN training can be up to one order of magnitude longer than ESN training. This is due to the combination of large values of $D$ and $M$.

Then, a trade-off could be considered in terms of optimization and training time. Exploring large values of $D$ and $M$, depending also on the input dimension $P$, can be costly ($\sim 51$h the longest case), whereas the optimization time for our ESN models took less time ($\sim 22$h the longest case). However, once the hyperparameters have been chosen for a given task, training and retraining over new trajectories could be much faster with the TN model. This could be important for tasks requiring quick adaptation and an online retraining scheme \cite{yan2024emerging}, such as when monitoring physical systems, since environmental conditions can change \cite{antonik2016online}. We note that we simplified the optimization of the ESN models by exploring just two hyperparameters: the spectral radius and the regularization constant. If more hyperparameters are included, the hyperparameter search could be even more expensive for ESNs than for TN models.

 In summary, we find that TN models can achieve performance similar to that of ESN models for short- and long-term metrics, while requiring at least one order of magnitude less training time. This makes them an interesting tool for time series prediction in industrial and large-scale applications.

\section*{Discussion}\label{sec:discussion}
Chaotic time series prediction is one of the most popular applications of RC architectures, but the selection and optimization of RC models is an open problem. In this work, we explore a TN model introduced by Batselier \cite{batselier2021enforcing,batselier2023khatri} that provides a low-rank approximation of the truncated Volterra series. Our goal is to provide a proof-of-concept for the suitability of TN models for time series prediction and, in particular, for learning chaotic systems. Our target tasks are a set of 70 standardized chaotic systems with polynomial nonlinearity, which serve as a perfect testbed. We benchmark the results of the TN model against those of a conventional ESN model to make a fair comparison with one of the most widely used models for predicting chaotic time series. We have shown a short-term memory metric and a climate metric to provide an indicator of the forecasting capacity of the TN model. We obtained a very similar performance between the TN and ESN models for most tasks with at least one order of magnitude smaller training time for the TN model. This, together with only two hyperparameters to optimize, makes the TN model very attractive for applications. 

This work is a first step in the exploration of TN models as state-space systems for learning chaotic systems, and as such, there is plenty of room for further exploration. A first step can be generalizing the TN implementation to predict complex spatiotemporal dynamics. Due to spatial discretization, the number of input features for these types of problems is usually of the order of hundreds or thousands of features, which, together with the need for a large number of training samples, demands a lot of computational resources for the SVD computation. A solution to this problem could be to compute a low-rank approximation of the randomized SVD \cite{batselier2018computing}. Another important research direction is an exhaustive comparison with the state-of-the-art models that populate the time series and RC literature, like Volterra and polynomial kernels \cite{gonon2022reservoir,grigoryeva2024infinite} and SINDy \cite{brunton2016discovering,kaptanoglu2023benchmarking}.To understand when each model can stand out, a systematic comparison must be carried out. Due to the scaling of their algorithms' time complexity, the number of training points and input features would define the suitability of each method. In principle, kernel methods are preferred for problems with many input features but not many data points, while state-space models should operate more efficiently for a few features and long datasets. Complex spatiotemporal tasks would be a promising benchmark because they require a large number of training points and input features. 

\section*{Appendix A: Metrics for all tasks} \label{ap:alltasks} 
\begin{figure*}[h!]
\captionsetup[subfigure]{}
\begin{center}
\includegraphics[scale=0.46]{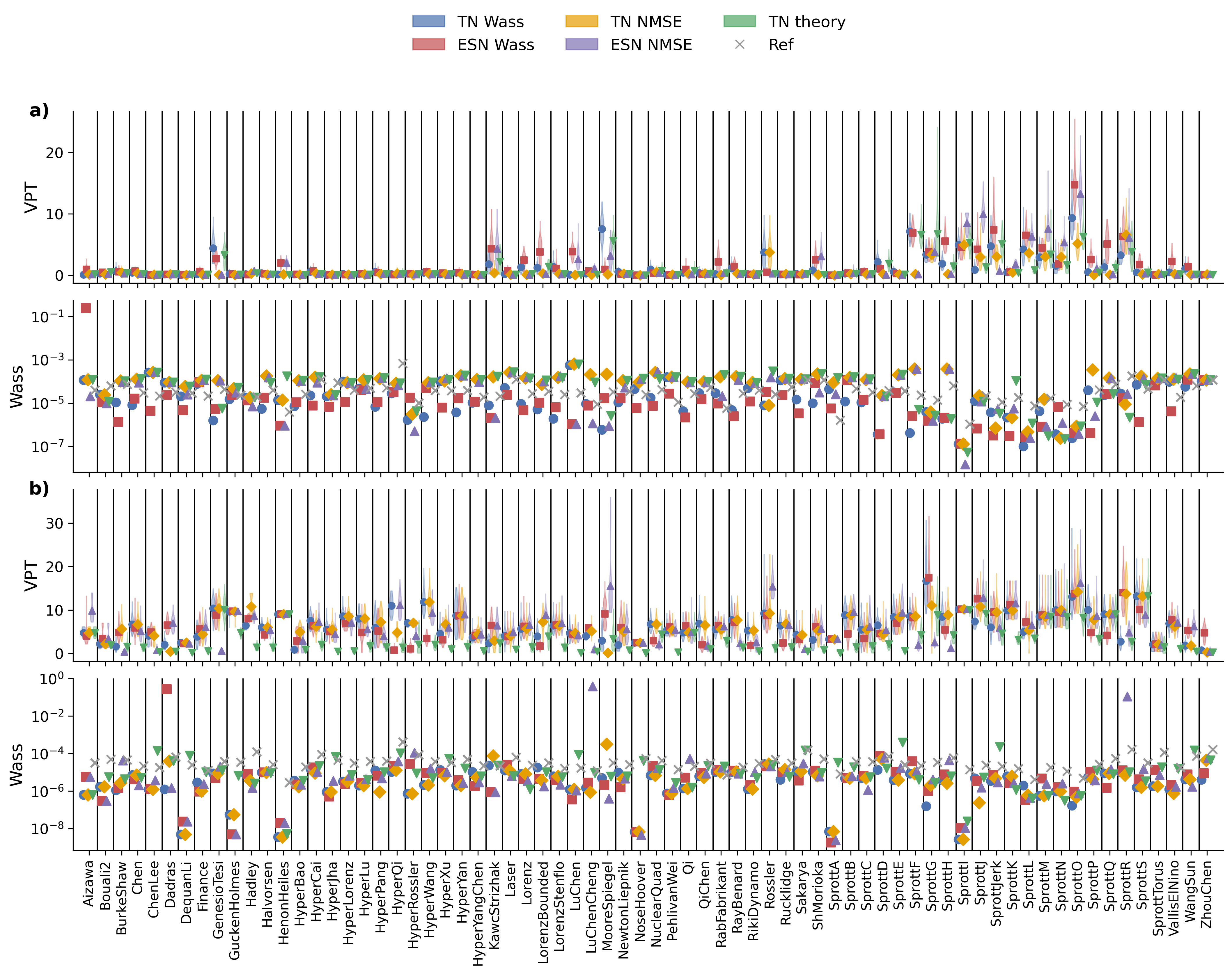}
\caption{Comparison of the VPT and Wasserstein metrics for the TN and ESN models for 70 tasks.}\label{fig3}
\end{center}
\end{figure*}

In this section, we break down the metrics for each of the 70 target tasks in Fig.~\ref{fig3}. The colour code is the same as in the main text, and the VPT results (in the first and third rows) are also represented with violins, but the points now represent mean values rather than median values. These mean values are averaged over 100 initial conditions. The reference values of the Wasserstein metric (gray crosses) are obtained by averaging over 50 trajectories with randomly generated initial conditions on the attractor.

The most eye-catching aspect of this figure is the difference between part of the Sprott family (Sprott F-S) and the rest of the tasks. This difference in performance is mildly appreciated in the memoryless case, while it is sharper in the memory case. We hypothesize that this is due to the small number of coefficients in the dynamical systems and their complexity, with only one quadratic coefficient and four to five linear terms. See the notebook \texttt{datasets.ipynb} in the source code for more details.

\section*{Appendix B: Minimal-norm and symmetric solution of $\textbf{H}$}\label{ap:Hsymm} 

Consider ${\bf U}$ an $N \times I^D$ matrix, where $N$ is the number of samples, $I=PM+1$ is the dimension of the extended input vector, $P$ is the input dimension, $M$ is the maximum delay and $D$ is the maximum order of monomials. 
The rows of ${\bf U}$ represent the vectorization of $D$th-order symmetric tensors $\mathcal{U}_n\in \R^{I\times \dots\times I}$ for $n=1,\dots,N$, that is,
\begin{equation}
    {\bf U}(n,:) = {\text{vec}}(\mathcal{U}_n), 
\end{equation}
where the symbol $(:)$ represents the whole range of index values. In this manuscript in particular, ${\bf U}(n,:) = {\bf u}^D_n$. Then, each row of ${\bf U}$ belongs to the vector space $\mathcal{S}_D$ of vectorized symmetric tensors of order $D$ in $\mathbb{R}^{I \times \cdots \times I}$, defined as
\begin{equation}
\begin{split}
\mathcal{S}_D = \{{\bf x}\in \R^{I^D} \mid  {\bf x}=\text{vec}(\mathcal{X}),\text{ } \mathcal{X}\in\text{Sym}_D(\R)\}.
    \end{split}
\end{equation}
Since every row ${\bf U}(n,:)$ lies in $\mathcal{S}_D$, the row space $\mathcal{R}({\bf U})$, which is the span of all the rows ${\bf U}(n,:)$, is a subspace of $\mathcal{S}_D$, that is, $\mathcal{R}({\bf U})\subset \mathcal{S}_D$, and any linear combination of the rows of ${\bf U}$ is a vectorization of a symmetric tensor. 

Let us now evaluate the minimal-norm solution of Eq.~\eqref{eq:Y=UH}. Consider a target matrix ${\bf Y}\in \R^{N\times L}$, the solution ${\bf H}\in\R^{I^D\times L}$ is given by the Moore–Penrose pseudoinverse:
\begin{equation}\label{eq:Hsym}
  {\bf H} = {\bf U}^+ {\bf Y} = {\bf U}^\top \left({\bf U} {\bf U}^\top\right)^+ {\bf Y} = {\bf U}^\top {\bf B},  
\end{equation}
where we used the identity $ {\bf U}^+={\bf U}^\top \left({\bf U} {\bf U}^\top\right)^+$ and defined the matrix factor $  {\bf B}:=\left( {\bf U}  {\bf U}^\top\right)^+  {\bf Y}$. Now we can see in Eq.~\eqref{eq:Hsym} that each column of ${\bf H}$ is formed as a linear combination of the columns of ${\bf U}^\top$, i.e. ${\bf H}(:,l) =  {\bf U}^\top {\bf B}(:,l)$ for $l=1,\dots L$. This corresponds to a linear combination of the rows of ${\bf U}$ when taking the transpose:
\begin{equation}
    {\bf H}(:,l)^\top= {\bf B}(:,l)^\top {\bf U} = \sum^N_{n=1}{\bf B}(n,l){\bf U}(n,:).
\end{equation}
Therefore, ${\bf H}(:,l)^\top\in \mathcal{R}({\bf U})\subset \mathcal{S}_D$. A very important detail is that we did not make any assumptions about the rank of ${\bf U}$, nor about the sizes of $N$ or $R$. Even if $\text{rank}({\bf U})<\min (N,R)$, the above discussion holds true. However, numerical computations are a different story. 

Next, we construct two examples to show the deviations from theory that numerical experiments reveal regarding the symmetry of ${\bf H}$. The first case is based on constructing ${\bf U}$ with random uncorrelated unidimensional inputs given by a uniform distribution $[0,1]$. In this way, we ensure $\text{rank}({\bf U})=\min (N,R)$. The second case is based on constructing ${\bf U}$ with highly correlated inputs given by the unidimensional sinusoidal signal
\begin{equation}
   u(n)=\frac{1}{2}\left(1+\sin\left(2\pi n\frac{2.11}{100}\right) \sin\left(2\pi n\frac{3.73}{100}\right)\sin\left(2\pi n\frac{4.11}{100}\right)\right),
\end{equation}
such that $\text{rank}({\bf U})<\min (N,R)$. Now we fix the hyperparameters $M=D=4$ and define a target for a simple task: we choose the memory task of recalling the last input, $\hat{y}(n) = u(n-1)$, and solve Eq.~\eqref{eq:Y=UH} by directly computing the pseudoinverse of ${\bf U}$ with the function $\texttt{pinv()}$ of \textit{Julia}. To check the symmetry of a $D$th-order tensor $\mathcal{H}\in \R^{I\times\cdots \times I}$, we compute the Euclidean distance between this tensor and all its permutations $\widehat{\mathcal{H}}_l$, for $l=1,\dots, D!$, where  $\widehat{\mathcal{H}}_l(j_1,\cdots, j_D)=\mathcal{H}(\pi_l(i_1,\cdots, i_D))$. The symmetry metric is then defined as
\begin{equation}
    S = \sum^{D!}_{l=1}\|\mathcal{H} - \widehat{\mathcal{H}}_l\|.
\end{equation} 
\begin{table}[h]
\begin{tabular}{|c|c|c|c|c|c|}
 \hline
 Input & $N$ & $R$ & $\text{rank}({\bf U})$ & $\text{rank}({\bf U}{\bf U}^\top)$ & $S$\\
 \hline
Uncorr. & 50 & 70 & 50 & 50 & $1.37\times 10^{-11}$ \\
Corr. & 50 & 70 & 48 & 41 & $0.13$ \\
Uncorr. & 100 & 70 & 70 & 70 & $8.80\times 10^{-12}$ \\
Corr. & 100 & 70 & 70 & 52 & $0.10$\\
 \hline
\end{tabular}
\caption{Numerical values of rank and $S$ for correlated and uncorrelated inputs.}
\label{table:svd}
\end{table}

Table \ref{table:svd} shows the values of $S$ for the single column of ${\bf H}$ together with the rank of ${\bf U}$, which is measured with the function $\texttt{rank()}$ of \textit{Julia} for both ${\bf U}$ and ${\bf U}{\bf U}^\top$. We can make two observations about these results. First, we note that for uncorrelated inputs, where $\text{rank}({\bf U})=\min (N,R)$,  the mathematical condition $\text{rank}({\bf U})=\text{rank}({\bf U}{\bf U}^\top)$ is fulfilled, while for correlated inputs, where $\text{rank}({\bf U})<\min (N,R)$,
different values of $\text{rank}({\bf U})$ and $\text{rank}({\bf U}{\bf U}^\top)$ are obtained. Then, there is a problem in the numerical computation of the function $\texttt{rank()}$ for correlated inputs. The second observation is that uncorrelated inputs yield a favorable symmetry metric ($S\sim 0$), while correlated inputs break the symmetry of ${\bf H}$. The latter happens even though, theoretically, the columns of ${\bf H}$ must be symmetric independently of how ${\bf U}$ is generated. Since we compute ${\bf H}$ through the pseudoinverse ${\bf U}^+$, the issue must be related to the numerical computation of $\texttt{pinv()}$. 

\begin{figure}[h!]
\captionsetup[subfigure]{}
\begin{center}
\includegraphics[scale=0.15]{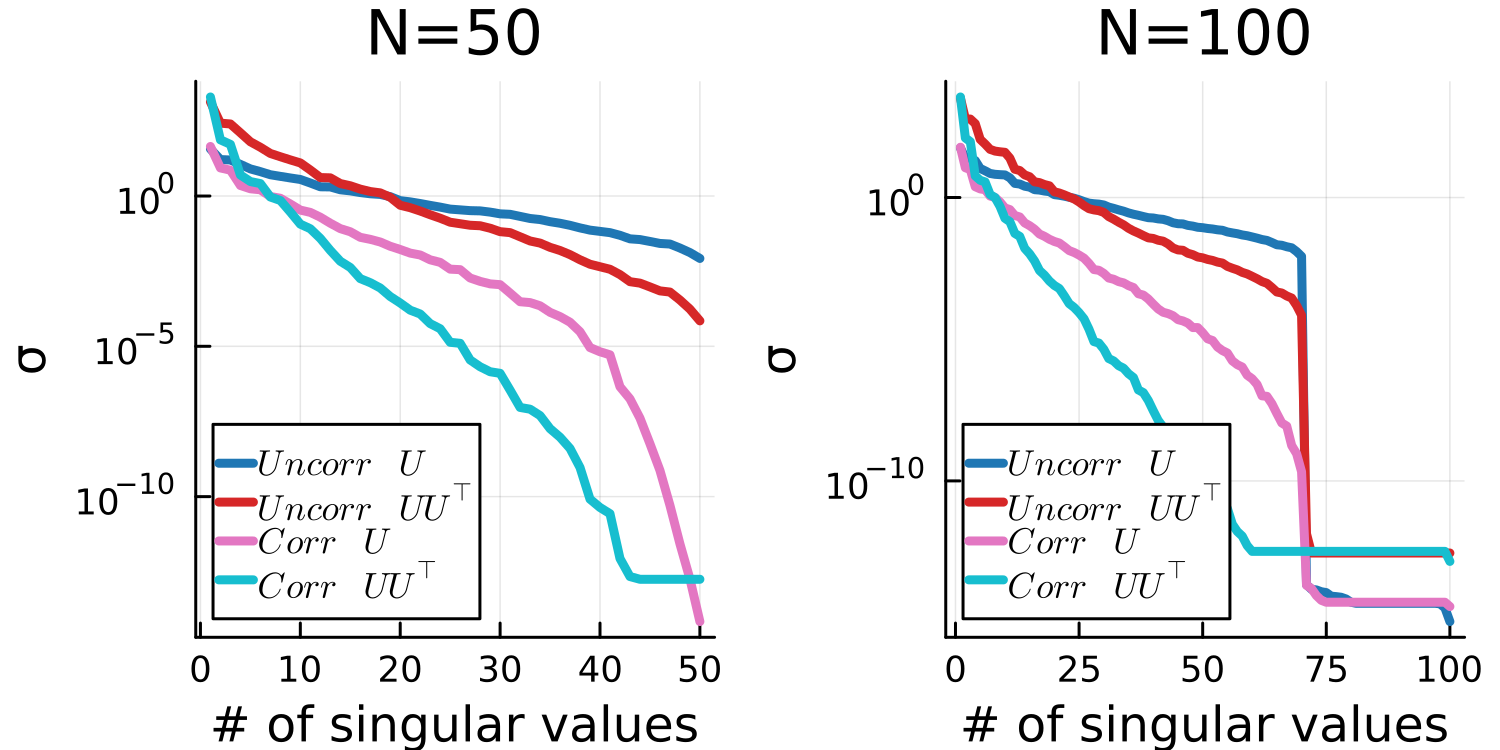}
\caption{Singular values of ${\bf U}$ for $N=50,100$ for $P=1$, $M=4$ and $D=4$ when given random uncorrelated inputs (blue) and correlated data (red) from the example of Table \ref{table:svd}. }\label{fig4}
\end{center}
\end{figure}

To understand these numerical deviations, we display the singular values of ${\bf U}$ and ${\bf UU}^\top$ from Table \ref{table:svd} in Fig.~\ref{fig4}. There, we observe that uncorrelated inputs (blue and red lines) produce singular values with a clear cutoff, while correlated inputs (magenta and cyan lines) generate a smooth slope in the magnitude of the singular values. Therefore, establishing a threshold for the truncation of singular values seems a harder problem for the case of generating ${\bf U}$ with correlated data. 

Indeed, the threshold problem explains our previous observations. Both the pseudoinverse computation and the rank estimation of ${\bf U}$ are based on the SVD computation and the truncation of the smaller singular values. As Eq.~\eqref{eq:Hsymm_solution} shows, the pseudoinverse is constructed by inverting the diagonal matrix of singular values, and those singular values that are very close to zero need to be removed to ensure the numerical stability of the computation. Then, the SVD computation
for ${\bf U}^+$ is much more stable for uncorrelated inputs than for correlated inputs, what is translated to the computation of $S$. The same idea applies to the deviations between $\text{rank}({\bf U})$ and $\text{rank}({\bf U}{\bf U}^\top)$, where the smoother behaviour of the singular values of ${\bf U}$ and ${\bf U}{\bf U}^\top$ yield discrepancies when setting a numerical threshold.

In summary, the symmetry of the columns of ${\bf H}$ is well preserved under $\text{rank}({\bf U})=\min (N,R)$, while the case $\text{rank}({\bf U})<\min (N,R)$ can break this symmetry because of numerical instabilities. In fact, careful attention must be paid to numerical experiments related to the SVD computation. Decisions like the choice of programming language can produce different results, since different languages use different criteria to truncate the singular values.

In this paper, we found that some of the chaotic tasks fulfill $\text{rank}({\bf U})<\min (N,R)$ due to the correlations between inputs at different time steps. In those cases, the functions $\text{rank}({\bf U})$ and $\text{rank}({\bf U}{\bf U}^\top)$ return different estimations, and we found a clear correlation: the higher is the discrepancy between estimations, the higher is $S$. See the notebook \texttt{test\_symmetry.ipynb} in the source code for the details. 

\section*{Data and code availability}

Our data and code are available via the following link: \url{https://github.com/RMPhys/tnrc_paper}

\section*{Acknowledgements}

We thank Hannah Lim Jing Ting, Miguel C. Soriano, and Juan-Pablo Ortega for their inspiration and useful discussions. We would like to thank Lyudmila Grigoryeva for their attentive reading and insightful
comments on this manuscript. RMP acknowledges the QCDI project funded by the Spanish Government.

\section*{Author contributions}

RMP and RO designed the research; RMP conducted the simulations and analyzed the results; both authors wrote and reviewed the manuscript.

\section*{Competing interests}

The authors declare no competing interests.

\end{document}